\documentclass{article}

\usepackage{tcolorbox}

\newtcolorbox{callout}{
    colback=gingerbeer!80,
    colframe=black,
    coltext=black,
    boxrule=1pt,
    arc=5pt,
    left=5pt,
    right=5pt,
    top=3pt,
    bottom=3pt,
}

\usepackage[preprint]{neurips_2026}              

\usepackage[utf8]{inputenc} 
\usepackage[T1]{fontenc}    
\usepackage{hyperref}       
\usepackage{url}            
\usepackage{booktabs}       
\usepackage{amsfonts}       
\usepackage{nicefrac}       
\usepackage{microtype}      
\usepackage{xcolor}         
\usepackage{graphicx}

\title{Are LLMs becoming similarly creative? Evidence from three years of models}

\author{%
  Nirav Patel$^{1}$\thanks{Corresponding author: \href{mailto:nirav.patel@duke.edu}{\texttt{nirav.patel@duke.edu}}} \quad
  Josiah Crossman$^{1}$ \quad
  Eva Aggarwal$^{1}$ \quad 
  Emily Wenger$^{1,2}$
  \\
  $^{1}$Department of Computer Science, Duke University \\
  $^{2}$Department of Electrical \& Computer Engineering, Duke University}

\begin{document}
\newcommand{\todo}[1]{\textcolor{red}{TODO: #1}}
\newcommand{\ejw}[1]{\textcolor{blue}{#1}}

\newcommand{\para}[1]{{\vspace{2pt} \bf \noindent #1}} 

\definecolor{copper}{HTML}{C84E00}
\definecolor{dandelion}{HTML}{FFD960}
\definecolor{piedmont}{HTML}{A1B70D}
\definecolor{eno}{HTML}{339898}
\definecolor{shale}{HTML}{0577B1}
\definecolor{dukeblue}{HTML}{012169}
\definecolor{ironweed}{HTML}{993399}
\definecolor{whisper}{HTML}{F3F2F1}
\definecolor{gingerbeer}{HTML}{FCF7E5}
\hypersetup{
  colorlinks=true,
  linkcolor=dukeblue!70!white,
  urlcolor=dukeblue!70!white,
  citecolor=dukeblue!70!white
}

\maketitle

\begin{abstract}
Many benchmarks track Large Language Model (LLM) performance on tasks with verifiable answers, but less is known about how LLM performance is evolving on open-ended tasks, where creativity, originality and diversity may matter as much as quality. As LLMs increasingly support human ideation and creative work, understanding trends in LLM performance on open-ended tasks is critical. This paper presents a preliminary analysis of LLM creative outputs spanning three years of model releases, examining model responses to Infinity-Chat100, a real-world collection of open-ended user queries, and the Alternate Uses Task, an established psychometric creativity assessment. Using sentence-embedding similarity, we examine trends in LLM responses to these prompts. Our findings show a statistically significant decrease in model output diversity over time, suggesting that LLM outputs may be converging in creative substance across models. If this trend persists, LLM-driven homogenization may progressively diminish human agency in human-AI co-creative work, demanding careful consideration of LLMs' role in the human creative process.
\end{abstract}

\section{Introduction}
\label{sec:intro}

In a few short years, conversational AI systems powered by large language models (LLMs) have moved from experimental prototypes to widely-used everyday tools. By July 2025, ChatGPT had more than 700 million weekly active users, who collectively sent more than 2.5 billion messages per day, or roughly 29,000 messages per second ~\cite{chatterji2025howpeopleusechatgpt}. Anthropic and DeepSeek, other popular AI chatbot providers, similarly have tens of millions of users~\cite{baptista2026deepseeklowcost, anthropic2025economicindex}. 

Studies show a wide variety of use cases for LLM-powered products, but creativity-relevant tasks like writing, content creation, and research are common. Approximately 70\% of consumer ChatGPT use is non-work-related, with practical guidance (a category which includes ``creative ideation'') and writing accounting for over half of ChatGPT messages~\cite{chatterji2025howpeopleusechatgpt}. Similarly, Anthropic estimates that nearly 20\% of Claude messages fall into categories like ``content creation'' or ``multidisciplinary academic research and writing.''~\cite{tamkinClioPrivacyPreservingInsights2024}. As LLMs mediate more creative work, it is  critical to understand how these systems affect the range of ideas users encounter and how model-generated outputs may shape downstream human creativity. 

A growing body of research suggests that LLM outputs can appear individually creative but are often collectively homogeneous\textemdash similar either to other ``creative'' content produced by the same LLM or content produced by other LLMs. In one study of AI-assisted creative writing, writers with access to AI-generated ideas produced stories that were judged as more creative and better written than stories from writers working alone. However, the AI-assisted stories were also more similar to one another than stories written without AI assistance~\cite{doshi2024generative}. Across real-world open-ended queries, models exhibit both intra-model repetition and inter-model convergence ~\cite{jiang2025artificialhivemind}. Human baseline studies on standardized divergent thinking tasks again surface this pattern, finding that LLM outputs have similar or higher creativity scores than humans on these tasks but are substantially less diverse~\cite{wenger2026homogeneouslycreative, wang2026divergentcreativity, bellemarepepin2026divergentcreativity}.  

Algorithmic monoculture has long been a subject of academic concern (e.g.~\cite{wu2025generative, kleinbergAlgorithmicMonocultureSocial2021}), and observed homogeneity across AI creative outputs represents yet another facet of this well-documented problem.  However, it remains unclear whether this homogeneity is a temporary byproduct of a still-developing technology or an inevitable\textemdash potentially compounding~\cite{shumailovCurseRecursionTraining2023}\textemdash feature of statistical language models. Plausible forces point in both directions. A growing body of academic work suggests that generative models trained on overlapping data and optimized toward similar objectives will organize concepts and semantic relationships in increasingly similar ways ~\cite{huh2024platonic, vuWhatHappensWhen2025, westBaseModelsBeat2025}, resulting in similar outputs. Yet, as models become more capable, they could produce more context-sensitive, stylistically distinct, and domain-specific responses, counteracting homogeneity~\cite{jainLLMOutputHomogenization2025}.

All these factors suggest an urgent need to understand the {\em trend of homogeneity in AI-generated creative outputs}, but no such analysis exists. All existing work on homogeneity captures a snapshot of specific models' behavior, rather than a longitudinal view of model performance on creative tasks. Nor can examination of existing benchmarks for AI models elucidate this trend. Most popular benchmarks for LLMs, such as MMLU~\cite{hendrycks2020measuring}, HELM~\cite{Bommasani2021FoundationModels}, SWE-Bench~\cite{jimenez2024swebench}, and Humanity's Last Exam~\cite{phan2025lastexam}, judge model performance against explicit criteria with verifiable answers. In contrast, studying creative outputs requires evaluating models on questions for which response correctness is less important than its novelty, diversity, or contextual sensitivity.

\para{Our contribution.} This paper fills this gap by performing a preliminary temporal analysis of model responses to open-ended prompts. We sample models across popular model providers and release periods from 2023 to the present, generate responses to two complementary sets of creative prompts, and compare outputs across models and generations. Our analysis combines the Alternate Uses Tasks'~\cite{guilford1978alternate} assessment of divergent thinking with open-ended prompts from Infinity-Chat100~\cite{jiang2025artificialhivemind}, which span six categories of real-world use, from creative content generation and alternative styles of writing, to information seeking, brainstorming and ideation. 

We then analyze trends in model response diversity by computing responses' semantic similarity via embedding-based distance metrics. This study provides a framework and initial findings for tracking how LLM creativity is evolving across model generations.

\para{Key findings.} We find that {\bf LLM responses to the open-ended prompts we test\textemdash AUT and Infinity-Chat100\textemdash have become increasingly similar over time}. This suggests that LLMs are becoming less creative in tasks that involve generating open-ended responses, demanding scrutiny of their long-term usefulness as creative assistants.

\section{Related Work}
\label{sec:back}

\para{Measuring creativity in LLMs.} Recent work studying creativity in LLMs adapts a family of methods derived from human psychometric assessments of creativity. These often include Torrance-style creativity dimensions such as fluency, flexibility, originality, and elaboration~\cite{torrance1966torrance}, as well as divergent-thinking tasks that measure novelty through semantic distance~\cite{zhao2025assessing, nakajima2026beyond}. Studies of AI creative writing combine expert human judgments, Consensual Assessment-style methods, and Torrance-inspired rubrics to test whether LLM outputs are genuinely creative products rather than merely fluent text ~\cite{chakrabarty2024art}. Other benchmarks shift from writing to problem solving, asking whether models can produce physically feasible, unconventional solutions to everyday challenges ~\cite{tian2024macgyver}. More recent work on scientific ideation similarly treats creativity as domain-specific idea generation~\cite{ruan2026evaluating}.

Although applying human psychometric methods to AI models is imperfect~\cite{stella2023using}, comparative studies suggest that LLMs can approach, and sometimes exceed, human performance on standard creativity tasks. For example, Hubert et al found that models score better than humans on standard creativity tasks, including the Divergent Association Test (DAT)~\cite{olsonNamingUnrelatedWords2021} generating more original and elaborate answers~\cite{hubert2024current}. However, a larger study of 9,198 humans and 215,542 LLM observations finds that humans are slightly more creative on average than LLMs on the DAT~\cite{wang2026large}. Recent work similarly finds that LLMs exceed average human DAT performance but fall below the most creative human subgroups. Notably, there was significant repetition across model DAT responses\textemdash “ocean” appeared in more than 90 percent of models' DAT response sets ~\cite{bellemarepepin2026divergentcreativity}. This suggests an emerging ``AI creativity paradox'': LLMs may appear individually creative but produce less diverse outputs in aggregate.

\para{Homogeneity in LLMs.} A growing body of literature documents this homogeneity across creative settings. In short-story generation, LLMs produce fluent and stylistically complex text, but score below humans on novelty, surprise, and diversity in expert and automated evaluations~\cite{ismayilzada2025evaluatingcreativeshortstory}. AI-assisted writing studies find a similar tradeoff, in which LLM-generated ideas improve individual story quality but reduce collective diversity~\cite{doshi2024generative}. Still more studies show that LLMs generate responses that are more similar to one another than human responses are, reuse plot elements, and exhibit both intra-model repetition and inter-model homogeneity across real-world open-ended prompts ~\cite{wenger2026homogeneouslycreative, jiang2025artificialhivemind, xu2025echoes}. 

While these observations of homogeneity are concerning, existing work merely provides snapshots of this phenomenon at a particular point in time. Furthermore, trends in homogeneity cannot easily be reverse-engineered from existing collections of benchmark data, which judge model performance against explicit criteria with verifiable answers (e.g.~\cite{hendrycks2020measuring,Bommasani2021FoundationModels,jimenez2024swebench,phan2025lastexam}). Even creativity-specific benchmarks, like CreativityPrism, LitBench, and Deep Associations, focus largely on whether LLMs can meet recognizable standards of creativity~\cite{hou2026creativityprismholisticevaluationframework, fein2026litbench, qiu2025deep}, not relationships between responses. Thus, no framework exists for tracking how the diversity or originality of LLM creative outputs is evolving.

\begin{figure}[t]
    \centering
    \includegraphics[width=\linewidth]{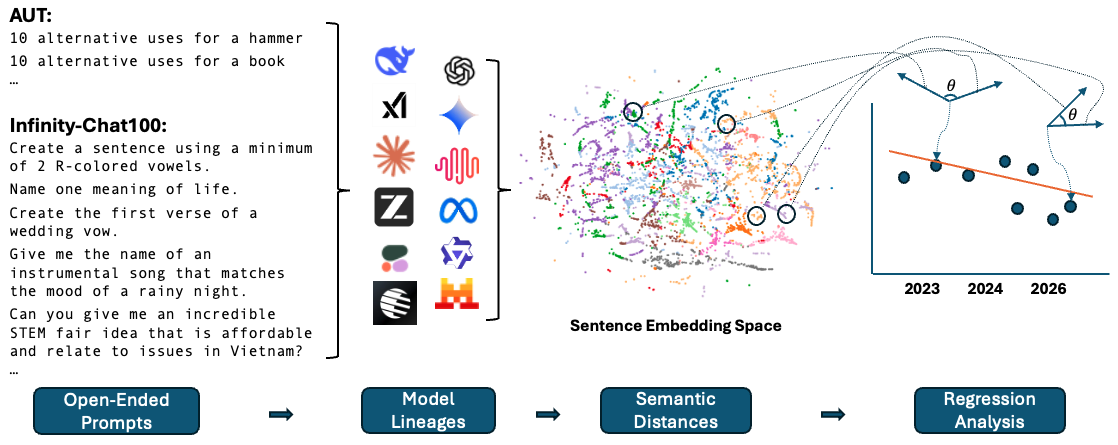}
    \caption{\em \small {\bf Methodology overview:} we select {\bf open-ended prompts} designed to elicit creative behavior from models; run these against different {\bf model lineages}; embed model outputs in a semantic space to compute their {\bf semantic distance}; and perform a {\bf regression analysis} to understand changes in semantic distance over time.}
    \label{fig:method_overview}
\end{figure}

\section{Methodology}
\label{sec:method}

We perform our time-based analysis of LLM output diversity via a four step process. We first select a set of open-ended prompts, run these across generations of language models, embed the responses in a semantic space to compute distances between them, and perform regression analysis to study trends in response distance. These steps are summarized in Figure~\ref{fig:method_overview} and described in detail below.

\textbf{Selecting prompts.} We select two prompt sets that elicit creativity in distinct ways: the Alternate Uses Task (AUT)~\cite{guilford1978alternate} and 100 real-world open-ended prompts from Infinity-Chat100~\cite{jiang2025artificialhivemind}. The AUT is a classic divergent-thinking assessment from cognitive psychology in which participants generate unconventional uses for common objects such as a book, shoe, or hammer. Its standardized structure isolates differences in the ideas models produce under consistent task constraints, providing a controlled setting for measuring responses' semantic similarity.

Conversely, the Infinity-Chat100 prompts provide a broader, more natural set of creative tasks. They were drawn from Infinity-Chat~\cite{jiang2025artificialhivemind}, a large dataset of real-world user interactions with language models, and span creative content generation, problem-solving, brainstorming, ideation, and other open-ended requests. These prompts permit substantial variation in both what models produce and how they approach each task. This naturally complements the more structured approach of the AUT. 

\textbf{Evaluating prompts against model lineages.} To construct the response dataset, we evaluate these prompts against 68 models released between March 2023 and July 2026, representing 12 major model providers. This covers 33 closed weight models and 34 open weight models across 27 release months (See ~\ref{sec:appendix} for model coverage details). Each model receives the complete set of AUT and Infinity-Chat100 prompts through OpenRouter’s API, with temperature and top-$p$ set to 1.0 for all generations (See ~\ref{sec:appendix} for data coverage details). These settings retain stochastic sampling without distorting the output distribution, allowing us to measure similarity among independently generated responses without narrowing models toward their highest-probability outputs. 

We assign each model a release date based on the month and year they were made publicly available, and order the resulting responses accordingly. Release date may not perfectly capture the combination of changes that distinguishes successive model generations, including newer training data, architectures, and post-training methods. However, we believe it serves as a consistent observable proxy for each model’s position in the broader progression of LLM development. Trends across release periods should therefore be interpreted as associations with broader model development patterns rather than as causal effects of time itself.

\para{Compute embeddings and semantic distances.} Next, we embed model responses using the \texttt{all-MiniLM-L6-v2} sentence-transformer model and compute semantic distances between these embeddings. The sentence-transformer represents each response as a dense vector in a semantic space, such that responses with similar meanings are represented by vectors pointing in similar directions. For the AUT, the complete set of uses generated in response to each object prompt was embedded as a single response, yielding 10 embeddings per model. For the Infinity-Chat dataset, each response was embedded separately, yielding 100 embeddings per model. 

\textbf{Regression specification.} Finally, we perform a linear regression of semantic distance between model responses over time (See ~\ref{sec:appendix} for OLS assumptions). To do this, we must first place the model pairs on a common timeline. We chronologically order the 27 ({\em month, year}) timestamps in our sample and group them into nine successive bins of three timestamps. Bins therefore span unequal calendar durations, because model releases have become more regular as more providers enter the market and existing providers iterate more rapidly. Grouping releases into equal-count bins ensures that each bin, even in the sparse early period, contributes measurable information to the linear regression. Using nine bins of three timestamps each balances a reasonable number of models per bin with appropriate separation between model releases. Following~\cite{wenger2026homogeneouslycreative}, we only compare responses between models from different providers (e.g. ``cross-family''), to minimize possible confounders like architecture or training data overlap. 

With our time bins established, we now compute distances between model responses within each bin. Formally, for each cross-family model pair within a bin, we match responses to the same prompts. Let $p \in \{1,\dots,P\}$ index prompts, with $P = 10$ for the AUT and $P = 100$ for Infinity-Chat100, and let $\mathbf{x}_{i,p}$ and $\mathbf{x}_{j,p}$ denote the embedding vectors produced by models $i$ and $j$ for prompt $p$. We compute prompt-level cosine distances as
\[
d_{ij,p} = 1 - \cos\left(\mathbf{x}_{i,p}, \mathbf{x}_{j,p}\right),
\]
where lower values indicate greater semantic similarity and higher values indicate greater divergence. We summarize each pair by its mean distance across prompts,
\[
\bar{d}_{ij} = \frac{1}{P}\sum_{p=1}^{P} d_{ij,p},
\]
which serves as the single observation for pair $(i,j)$, computed separately for each dataset. To test whether cross-family divergence changes systematically over time, we fit an ordinary least squares regression of $\bar{d}_{ij}$ on $b_{ij} \in \{1,\dots,9\}$, the ordinal index of the release bin containing the pair,
\[
\bar{d}_{ij} = \alpha + \beta\, b_{ij} + \varepsilon_{ij},
\]
so that $\hat{\beta}$ is interpretable as the average change in cross-family cosine distance per release bin.

\noindent {\em Model resampling for regression.} Because model families contain different numbers of models, an unadjusted regression would weight heavily represented families more strongly. To reduce this influence, we run 1{,}000 iterations of family-balanced resampling. At each iteration $k \in \{1,\dots,1000\}$, we sample one model from each family represented in each release bin, compute  distances between cross-family response sets from the sampled models, and perform a separate regression for each response set across all bins. This yields a distribution of slope estimates $\{\hat{\beta}^{(k)}\}_{k=1}^{1000}$. 

\begin{figure}[t]
    \centering
    \includegraphics[width=\linewidth]{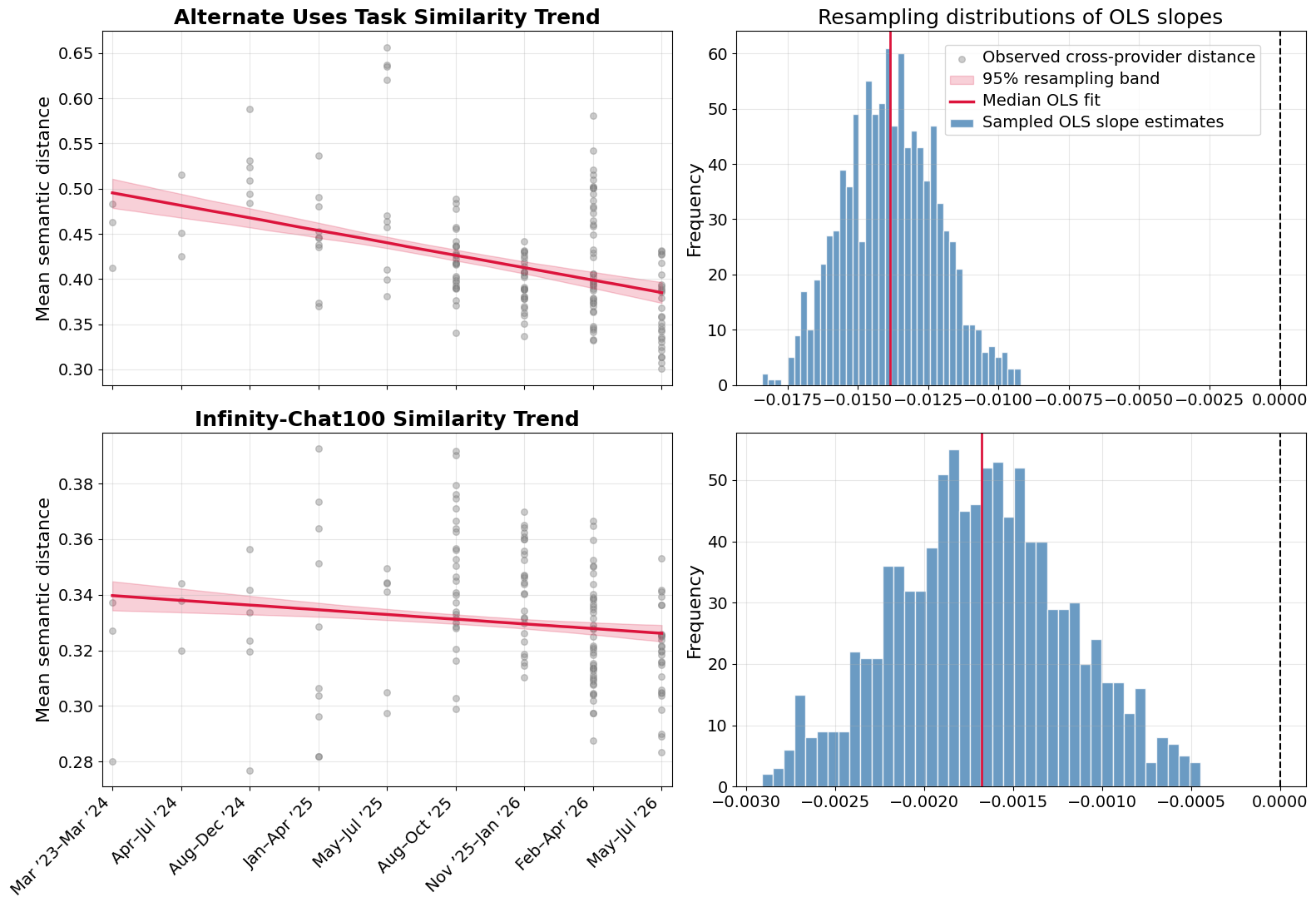}
    \caption{\small \em {\bf Regression results and bootstrapped slope estimates for AUT and Infinity-Chat response distances across model generations.} Our findings of consistently negative slopes across the observation period indicate semantic convergence across LLM creative outputs over time.}
    \label{fig:cosine_distance}
\end{figure}

 We summarize the temporal trend using the median slope across iterations, with the 2.5th and 97.5th percentiles reported as empirical resampling intervals. We additionally construct a pointwise 95\% band from the corresponding percentiles of the fitted regression lines at each bin index, which we display alongside the binned means in Figure~\ref{fig:cosine_distance}. These intervals quantify how sensitive the trend is to which model represents each family, addressing the confounding influence of unequal family representation at any point in time. Because our interest is not in any individual model but rather in the collective outputs of models in a given period, the model that happens to represent its family within a bin should not matter. A tight interval thus indicates that the trend is not an artifact of any particular selection of models and holds regardless of which model stands in for its family in each bin. A consistent negative slope then indicates decreasing cross-family divergence over time. 

\section{Results}
\label{sec:results}

\begin{callout}
{\bf Key finding:} For both the AUT and Infinity-Chat response sets, we observe a decline in cross-provider output distances over the observation period. This suggest decreasing diversity\textemdash or increasing homogeneity\textemdash of LLM creative outputs over time.
\end{callout}

Figure~\ref{fig:cosine_distance} shows that cross-family cosine distance declines over the observation period for both prompt sets, with the corresponding slope estimates reported in Table~\ref{tab:cosine_slope_stats}. The decline is most pronounced for the Alternate Uses Task, where mean cross-family distance falls from approximately 0.50 in the earliest release bin to below 0.40 in the most recent. Infinity-Chat exhibits the same pattern at a much gentler rate, declining from approximately 0.34 to just above 0.32. Notably, all 1{,}000 resampling iterations produced negative slopes for both prompt sets, indicating that the direction of the relationship is robust across the sampled combinations of models from each family.

\begin{table}[t]
    \centering
    \caption{\small \em Bootstrap OLS estimates for mean cosine distance across nine model release bins.}
    \label{tab:cosine_slope_stats}
    \begin{tabular}{lcccc}
        \toprule
        Dataset & Models (Pairs) & Slope per Bin & 95\% CI\\
        \midrule
        Alternate Uses Task & 68 (273) & $-0.01385$ & $[-0.01695, -0.01044]$\\
        Infinity-Chat        & 67 (268) & $-0.00167$ & $[-0.00267, -0.00074]$\\
        \bottomrule
    \end{tabular}
\end{table}

The magnitude of the AUT decline is particularly notable given the task’s purpose. The AUT explicitly tests divergent thinking by instructing models to produce uses that are as original and unexpected as possible, making it precisely the setting in which outputs would be expected to differ. 

The Infinity-Chat results extend this finding beyond a single, highly structured task format. Its 100 prompts elicit many forms of open-ended generation, ranging from imaginative and expressive tasks to reflective and conversational ones. A decline across this heterogeneous collection is therefore less likely to reflect the creative demands of any one task and instead suggests a broader change in the distinctiveness of models’ creative outputs over time. Although the overall Infinity-Chat slope is shallower, the decline is most visible in the recent release bins (e.g. 2025 onward). This pattern may mark the beginning of a longer-term trend that warrants continued monitoring.

\section{Discussion}

Our findings, though preliminary, raise concerns about the long-term usefulness of LLMs as creative partners. Even if models perform well on creative tasks, converging outputs could bound the range of possibilities LLM users are exposed to, and with it, the breadth of their own thinking. If using an LLM for creative tasks like essay writing decreases one's brain activity~\cite{kosmynaYourBrainChatGPT2025}, could using increasingly less creative LLMs\textemdash the trend suggested by our study\textemdash further worsen LLMs' effects on human creativity, as observed by this and other studies~\cite{doshi2024generative, Anderson_2024}? Future work should examine this possibility while accounting for the limitations of this initial work.

\para{Limitations.} Although our regression approach is rigorous, it cannot account for hidden relationships across model families, such as distillation~\cite{oai_api_distill, anthropic_distill}, shared training data~\cite{vuWhatHappensWhen2025}, or common training techniques. Second, we anecdotally observe several prompts in the InfinityChat dataset that request similar outputs (e.g. 2 prompts asking about the movie Zootopia). While this does not affect our measurement of response similarity, since we compute distances between model responses on the {\em same} prompts, it suggests the need for a more principled set of unstructured creative prompts. Moreover, while sampling a single response per model per prompt is sufficient for our aggregate trend analysis, it yields only a point estimate of each model's output distribution. Comparing distributions drawn from repeated generations could more richly characterize models' output creativity. Finally, our analysis does not account for the role of the user in shaping the diversity of LLM outputs. Recent work suggests that prompting alone cannot prevent homogeneity in AI creative outputs~\cite{wenger2026homogeneouslycreative}, but savvier use of AI models by humans may produce better results~\cite{urban2025prompting, joIncentivesShapeHow2026}.

\para{Future work.} Our work leaves open many avenues for interesting future work. Our aggregate analysis does not establish whether the observed creative similarity is uniform across all types of open-ended prompts, so a prompt-level decomposition may help identify the factors driving model convergence. For example, dividing prompts by task type, category, and output length could reveal whether similarity patterns differ across short-form creativity, long-form writing, or practical ideation. Our embedding-based distance measurements could be complemented by text-level and structural measures, such as Jaccard similarity, sentence structure, response organization, and other stylistic features. Further robustness checks across prompt wording, sampling parameters, response length constraints, model family, and model size would also help rule out confounders. Such directions would help clarify not only the existence of defined trends in LLMs' creative evolution, but also the underlying mechanisms driving them.

\newpage

\bibliographystyle{unsrt}
\bibliography{references}

\newpage
\appendix
\section{Appendix}
\label{sec:appendix}

\renewcommand{\thefigure}{A\arabic{figure}}
\setcounter{figure}{0}
\renewcommand{\thetable}{A\arabic{table}}
\setcounter{table}{0}

\begin{figure}[h]
    \centering
    \includegraphics[width=\linewidth]{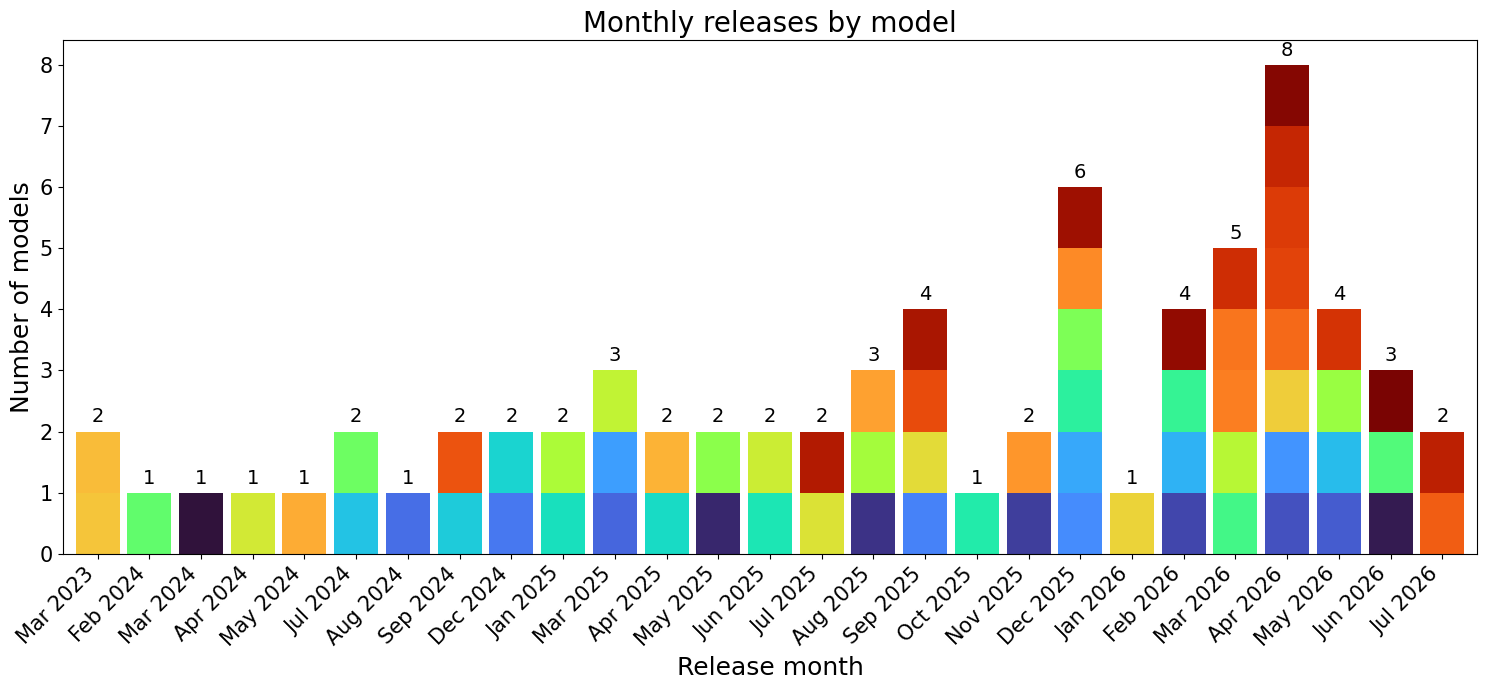}
    \vspace{-0.5em}
    \includegraphics[width=\linewidth]{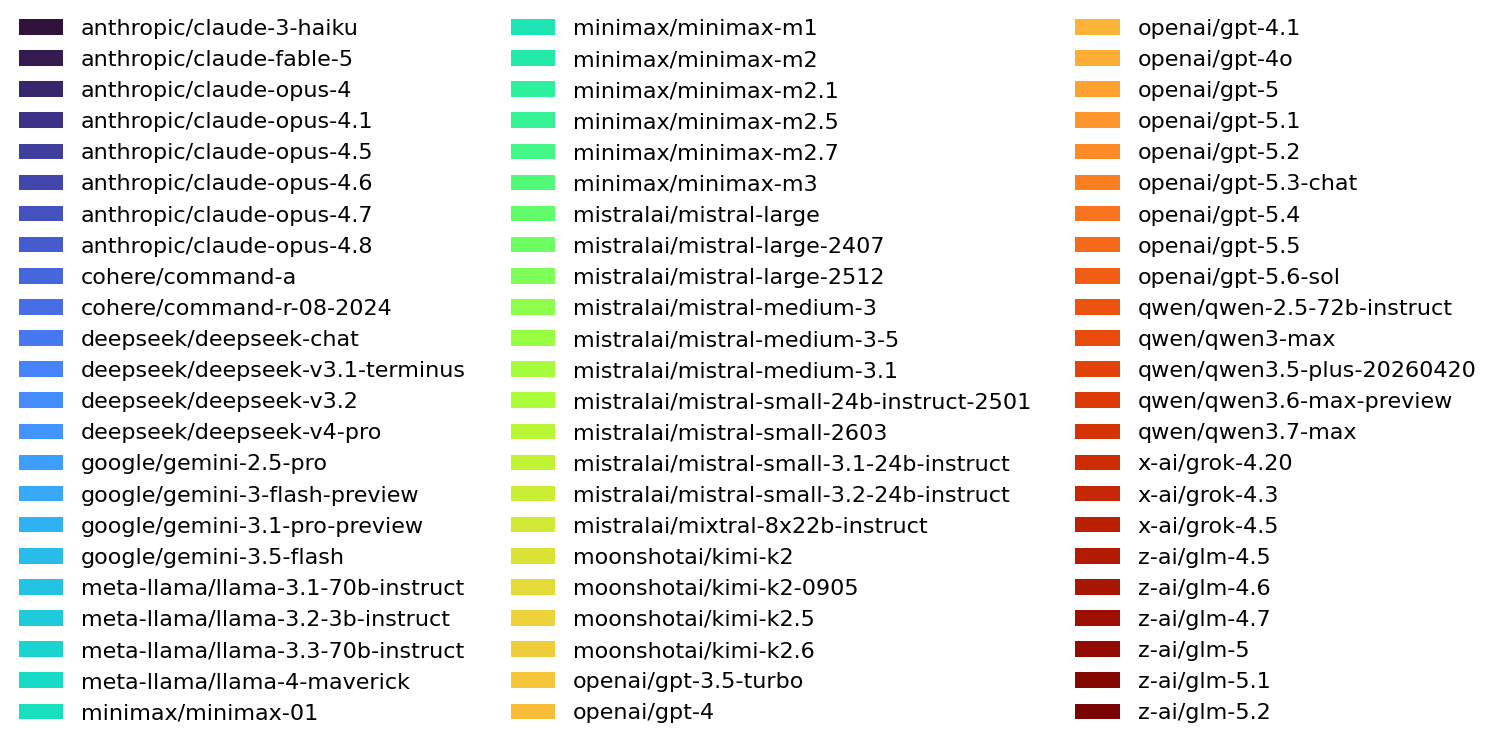}
    \caption{Distribution of model releases over time across the 12 model providers and 68 versions in our dataset.}
    \label{fig:model_time}
\end{figure}

\textbf{Model coverage details.} Figure ~\ref{fig:model_time} shows all the models used in our analysis along with their public release dates.

\textbf{Data coverage details.} Across 69 models, the AUT dataset contained 690 possible model-level prompt responses, of which 668 were observed and 22 were missing or dropped (3.19\%). Infinity-Chat100 contained 6,900 possible responses, of which 6,677 were observed and 223 were missing or dropped (3.23\%). Nine models had at least one missing response across the two datasets, while the remaining 60 models were complete in both. Table~\ref{tab:missing-responses} reports the observed and missing counts for these nine models. Note that responses are missing or dropped only when an API request failed after repeated trials or when the returned output was corrupted or unusable as a direct response to the prompt. Additionally, when coverage was incomplete, the mean cosine distance for each model pair was computed only across prompts for which both models had valid responses. This preserved the available pairwise information without imputing missing outputs. Given the low overall rate of missingness and aggregation across 1,000 model-sampling iterations, missing responses were unlikely to materially influence our reported estimates.

\begin{figure}[h]
    \centering
    \includegraphics[width=\linewidth]{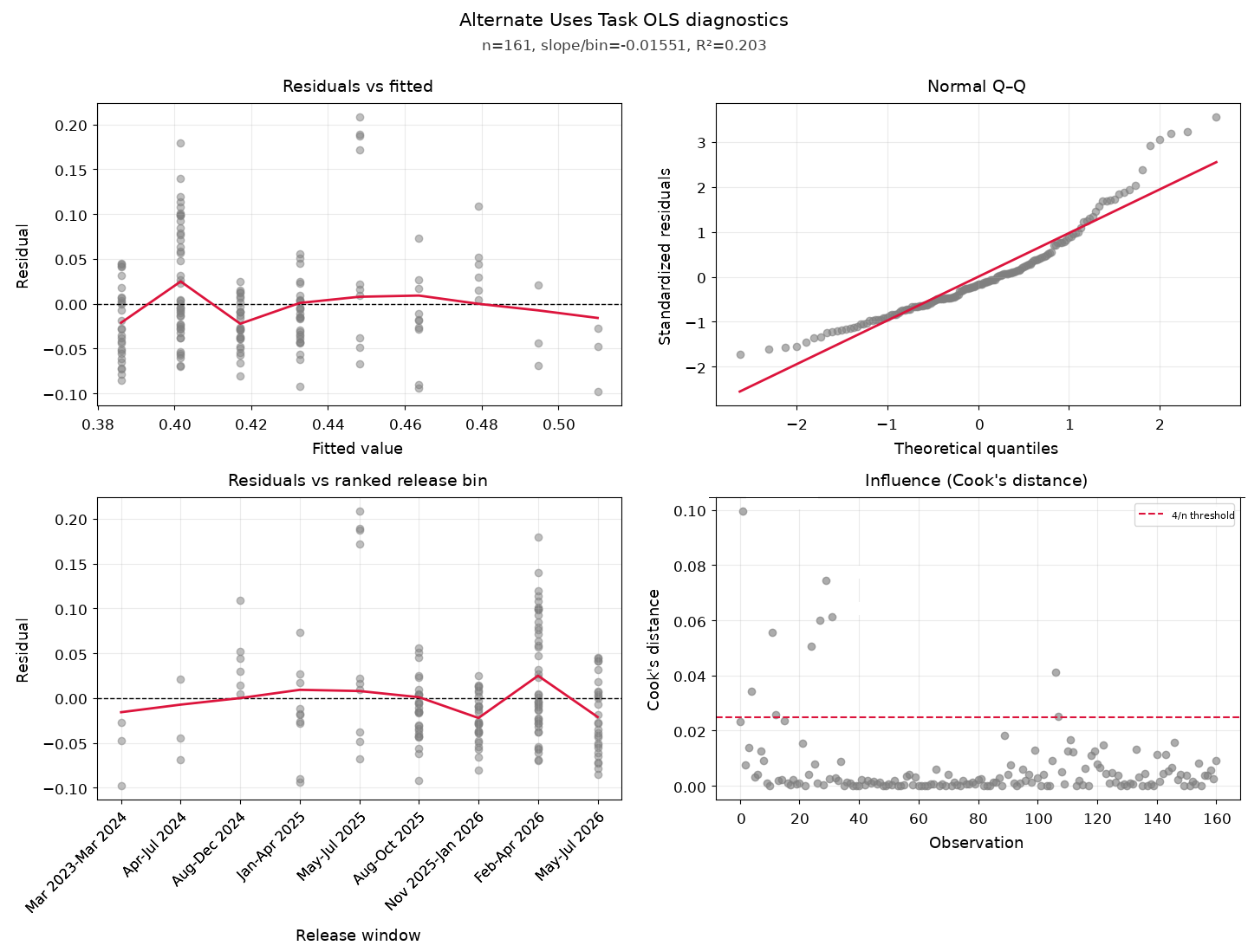}
    \vspace{-0.5em}
    \includegraphics[width=\linewidth]{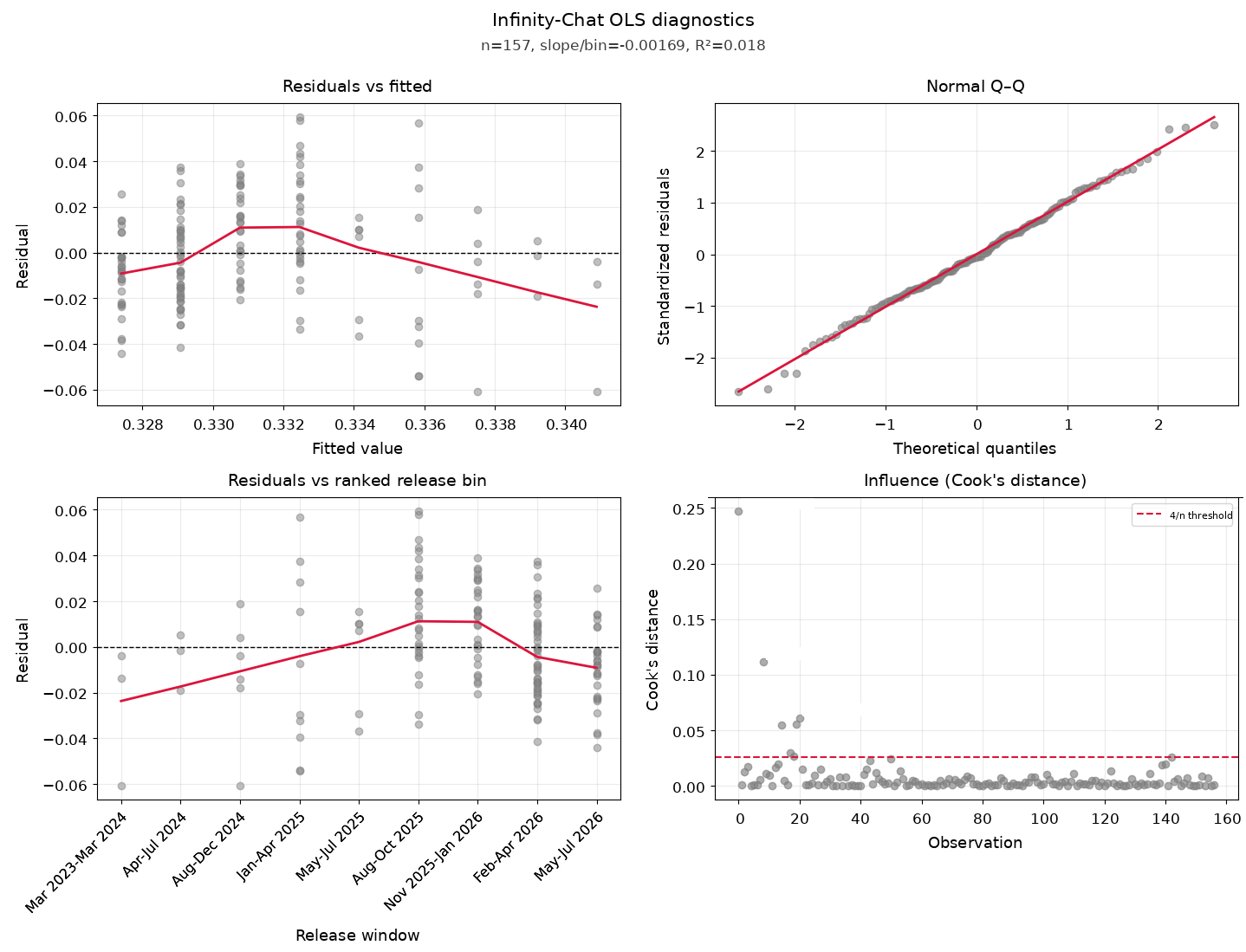}
    \caption{OLS Linearity, Normality, Independence, and Influential Observations}
    \label{fig:OLS_assumptions}
\end{figure}

\textbf{OLS assumptions.} Figure~\ref{fig:OLS_assumptions} reports standard OLS diagnostics for both regressions. Residuals are roughly centered across fitted values and release bins, and departures from linearity roughly track the number of models in each bin while the overall relationship remains approximately linear. The Q-Q plots indicate approximately normal residuals for Infinity-Chat and modest right-tail deviation for the AUT. The most influential observations by Cook's distance are concentrated in the earliest release bins, where fewer models were available and each cross-family pair carries more weight. Our family-balanced resampling further mitigates this by ensuring no single model dominates the estimates. As such, we do not expect these deviations to affect our analysis, but rather see them as confirming the need to continue collecting data as models continue to release.

\begin{table*}[h]
\caption{Observed and missing responses for models with at least one missing response. Each model had 10 possible AUT responses and 100 possible Infinity-Chat100 responses.}
\label{tab:missing-responses}
\centering
\small
\begin{tabular}{lrrrr}
\toprule
& \multicolumn{2}{c}{AUT}
& \multicolumn{2}{c}{Infinity-Chat100} \\
\cmidrule(lr){2-3}
\cmidrule(lr){4-5}
Model & Observed & Missing & Observed & Missing \\
\midrule
\texttt{minimax/minimax-01}
    & 0  & 10 & 0   & 100 \\
\texttt{minimax/minimax-m1}
    & 6  & 4  & 0   & 100 \\
\texttt{mistralai/mistral-small-3.1-24b-instruct}
    & 10 & 0  & 89  & 11 \\
\texttt{qwen/qwen3.6-max-preview}
    & 2  & 8  & 100 & 0 \\
\texttt{qwen/qwen3-max}
    & 10 & 0  & 94  & 6 \\
\texttt{meta-llama/llama-3.2-3b-instruct}
    & 10 & 0  & 97  & 3 \\
\texttt{anthropic/claude-fable-5}
    & 10 & 0  & 99  & 1 \\
\texttt{minimax/minimax-m2.1}
    & 10 & 0  & 99  & 1 \\
\texttt{qwen/qwen-2.5-72b-instruct}
    & 10 & 0  & 99  & 1 \\
\midrule
\textbf{All 69 models}
    & \textbf{668}
    & \textbf{22}
    & \textbf{6,677}
    & \textbf{223} \\
\bottomrule
\end{tabular}
\end{table*}

\end{document}